\documentclass[conference]{IEEEtran}
\IEEEoverridecommandlockouts
\usepackage{cite}
\usepackage{amsmath,amssymb,amsfonts}
\usepackage{algorithmic}
\usepackage{graphicx}
\usepackage{textcomp}
\usepackage{xcolor}
\usepackage{hyperref}
\hypersetup{
    colorlinks=false,
    pdfborder={0 0 0},
}
\def\BibTeX{{\rm B\kern-.05em{\sc i\kern-.025em b}\kern-.08em
    T\kern-.1667em\lower.7ex\hbox{E}\kern-.125emX}}

\usepackage[utf8]{inputenc}
\usepackage{textgreek} 

\begin{document}

\makeatletter
\def\ps@IEEEtitlepagestyle{
    \def\@oddfoot{\parbox{\textwidth}{\centering Accepted at the Sixth IEEE International Conference on Image Processing Applications and Systems (IPAS) 2025. }}
    \def\@evenfoot{}
}
\makeatother

\title{Rapid Distributed Fine-tuning of a Segmentation Model Onboard Satellites\\
}

\author{\IEEEauthorblockN{Meghan Plumridge}
\IEEEauthorblockA{
\textit{University of Cambridge}\\
Cambridge, United Kingdom \\
map205@cam.ac.uk}
\and
\IEEEauthorblockN{Rasmus Maråk}
\IEEEauthorblockA{\textit{AI Sweden} \\
Gothenburg, Sweden \\
rasmus.marak@ai.se }
\and
\IEEEauthorblockN{Chiara Ceccobello}
\IEEEauthorblockA{\textit{AI Sweden} \\
Gothenburg, Sweden \\}
\and
\IEEEauthorblockN{Pablo Gómez}
\IEEEauthorblockA{
\textit{European Space Astronomy Centre}\\
\textit{European Space Agency}\\
Madrid, Spain \\}
\and
\IEEEauthorblockN{Gabriele Meoni}
\IEEEauthorblockA{
\textit{European Centre for Earth Observation}\\
\textit{European Space Agency}\\
Frascati, Italy \\}
\and
\IEEEauthorblockN{Filip Svoboda}
\IEEEauthorblockA{\textit{University of Cambridge}\\
Cambridge, United Kingdom \\}
\and
\IEEEauthorblockN{Nicholas Lane}
\IEEEauthorblockA{\textit{University of Cambridge}\\
Cambridge, United Kingdom \\}
}

\maketitle

\begin{abstract}
Segmentation of Earth observation (EO) satellite data is critical for natural hazard analysis and disaster response. However, processing EO data at ground stations introduces delays due to data transmission bottlenecks and communication windows. Using segmentation models capable of near-real-time data analysis onboard satellites can therefore improve response times. This study presents a proof-of-concept using MobileSAM, a lightweight, pre-trained segmentation model, onboard Unibap iX10-100 satellite hardware. We demonstrate the segmentation of water bodies from Sentinel-2 satellite imagery and integrate MobileSAM with PASEOS, an open-source Python module that simulates satellite operations. This integration allows us to evaluate MobileSAM's performance under simulated conditions of a satellite constellation. Our research investigates the potential of fine-tuning MobileSAM in a decentralised way onboard multiple satellites in rapid response to a disaster. Our findings show that MobileSAM can be rapidly fine-tuned and benefits from decentralised learning, considering the constraints imposed by the simulated orbital environment. We observe improvements in segmentation performance with minimal training data and fast fine-tuning when satellites frequently communicate model updates. This study contributes to the field of onboard AI by emphasising the benefits of decentralized learning and fine-tuning pre-trained models for rapid response scenarios. Our work builds on recent related research at a critical time; as extreme weather events increase in frequency and magnitude, rapid response with onboard data analysis is essential. 
\end{abstract}


\section{Introduction}
Analysis of Earth observation (EO) satellite data is essential for disaster response. However, satellite data is not available to end-users in real-time; data must be down-linked from a satellite to a ground station and subsequently processed, before being distributed to end-users for further analysis. This workflow significantly delays the response to real-time hazards, as the satellite must first travel to a specific point before transmission to a ground station can begin. The situation is further complicated as the number of EO satellites deployed in orbit is projected to triple within the next decade \cite{Novaspace2024}, and these satellites are equipped with increasingly sophisticated instruments, generating unprecedented volumes of data. Transmitting these larger volumes of data to ground stations becomes increasingly challenging due to existing limitations in bandwidth and scarce communication windows. 

Recent advancements in satellite hardware offer promising solutions for improving rapid response capabilities \cite{furano2020towards}, \cite{Kothari2020}. Satellites equipped with central, graphical and vision processing units (CPUs, GPUs and VPUs) are now capable of running demanding onboard applications, reducing the reliance on ground-based data processing facilities. Important milestones have occurred within the last five years; the European Space Agency demonstrated the first instance of a deep learning model running on a commercial processor, in-orbit, as part of the PhiSat-1 mission \cite{phisat}. More recently, in 2023, \cite{vit} demonstrated the first instance of training a deep learning model onboard EO satellite hardware, highlighting the advantages of deploying pre-trained models onboard.

In addition to the adaptation of pre-trained models, distributing ML workflows also appears to be advantageous in terms of rapid response. Federated learning (FL) brings complex models to the data, avoiding the need for transmission of large volumes of data. The learning can be distributed among a number of agents within the distributed system that communicate and share their model updates via a central server. Constellations of satellites are suitable contenders for distributed workflows and very recent work has started to explore this topic. In the context of EO, FL has recently been demonstrated for scene classification and compression tasks \cite{ostman2023decentralised,gomez2024}. Work by \cite{Razmi2023} also provides an overview of FL in the context of EO missions.

In this study, we further explore the topic of fine-tuning pre-trained models onboard EO satellites, in a decentralised way, to support rapid response during disaster scenarios by:
\begin{enumerate}
    \item Fine-tuning a pre-trained segmentation model on a satellite processor, to minimise the amount of time and data required for model fine-tuning.
    \item Distributing fine-tuning between multiple satellites within a constellation, to explore how the rate of fine-tuning varies.
\end{enumerate}

\subsection{Segmentation Modelling}
Segmentation models are effective at hazard geo-location. A review of segmentation methods by \cite{Bagwari_Kumar_Verma_2023} concluded that deep learning, in particular convolutional neural networks (CNNs), are most promising for EO data, requiring less manual intervention compared with traditional methods. However, despite the adoption of CNNs to process payload data onboard satellites \cite{Meoni2024, Mateo-Garcia_Veitch-Michaelis_Smith_Oprea_Schumann_Gal_Baydin_Backes_2021, 9600851}, their use for scenarios with scarce labeled training examples is limited \cite{Meoni2024, sebastianelli2021automatic}. Our study aims to address this problem by fine-tuning a lightweight version of Meta AI's segment anything model (SAM) \cite{Kirillov_Mintun_Ravi_Mao_Rolland_Gustafson_Xiao_Whitehead_Berg_Lo_2023}. 

SAM has been trained on more than 1 billion images and is regarded as a benchmark model for computer vision. It has demonstrated exceptional generalisation capabilities, requiring little to no re-training when working with unseen datasets. A study by \cite{Osco_Wu_deLemos_Gonçalves_Ramos_Li_Junior_2023} investigated the applicability of SAM for remote sensing applications. Despite some promising results, the authors recommend exploring fine-tuning to improve performance with coarser resolution EO data, which we do in this study. Recently, several light-weight implementations of SAM have been developed, which opens up the possibility of deploying these lighter weight models onboard satellite hardware. We use MobileSAM, which is 60 times smaller than the original SAM, due to the use of a smaller image encoder, with comparable performance \cite{Zhang_Han_Qiao_Kim_Bae_Lee_Hong_2023}. By using a foundation  model, we significantly limit the need for labelled training data. Moreover, by moving the training directly onboard, fine-tuning can be performed as soon as data are acquired. 

\subsection{Distributed learning with PASEOS}
Training models in space poses fundamentally different constraints than on Earth. In orbit, spacecraft need to regulate their temperature, manage power efficiently, and deal with constantly changing availability and limited bandwidth of communication links. This influences the feasibility of running ML methods onboard a satellite. One way to simulate the operational environment of EO satellites is with the open-source Python module PASEOS \cite{Gómez_Östman_Shreenath_Meoni_2023}. PASEOS offers utilities for monitoring onboard and operational constraints, including thermal, power, bandwidth and communication constraints. In this work, we use PASEOS to observe power and temperature trends during training of MobileSAM onboard multiple satellites, as well as communication windows. We demonstrate the segmentation workflow on a Unibap iX10-100 satellite processor for simulated operational environments.

\subsection{Contributions}
The key contributions of this work are as follows:
\begin{enumerate}
    \item We demonstrate for the first time use of a SAM model onboard satellite hardware.
    \item  We explore different distributed learning scenarios among EO satellites within a constellation, with a key goal of rapid response. 
    \item We integrate MobileSAM with the PASEOS simulation software, developing a code base that enables easy reproduction of our results.
    \item We demonstrate reduced latencies, from initial observation to actionable information of disaster scenarios.
\end{enumerate} 
This study explores these concepts whilst considering typical operational constraints for onboard satellite constellation scenarios. Our demonstration is also relevant for the planetary science community, where bandwidth bottlenecks are significantly worse compared to Earth orbit-based scenarios.  

\section{Methodology}
Our methodology consists of: 1) data acquisition and pre-processing, 2) integration of MobileSAM with PASEOS, 3) model deployment and bench-marking on satellite hardware.

\subsection{Data Selection and Preparation}

The WorldFloods dataset \cite{portales-julia_global_2023, Mateo-Garcia_Veitch-Michaelis_Smith_Oprea_Schumann_Gal_Baydin_Backes_2021} is used for model fine-tuning. The dataset comprises Sentinel-2 L1C satellite images and corresponding flood segmentation masks for global flood events between 2019 and 2023 \cite{Mateo-Garcia_Veitch-Michaelis_Smith_Oprea_Schumann_Gal_Baydin_Backes_2021, portales-julia_global_2023}. The complete dataset is 315 GB in size and already divided into training, validation and testing subsets, ensuring diverse global sampling. We process the data according to the steps below; an example satellite image and ground truth mask is shown in Fig. \ref{fig:TrainingData}: 
\begin{figure}[!b]
    \centering
    \includegraphics[width=.99\columnwidth]{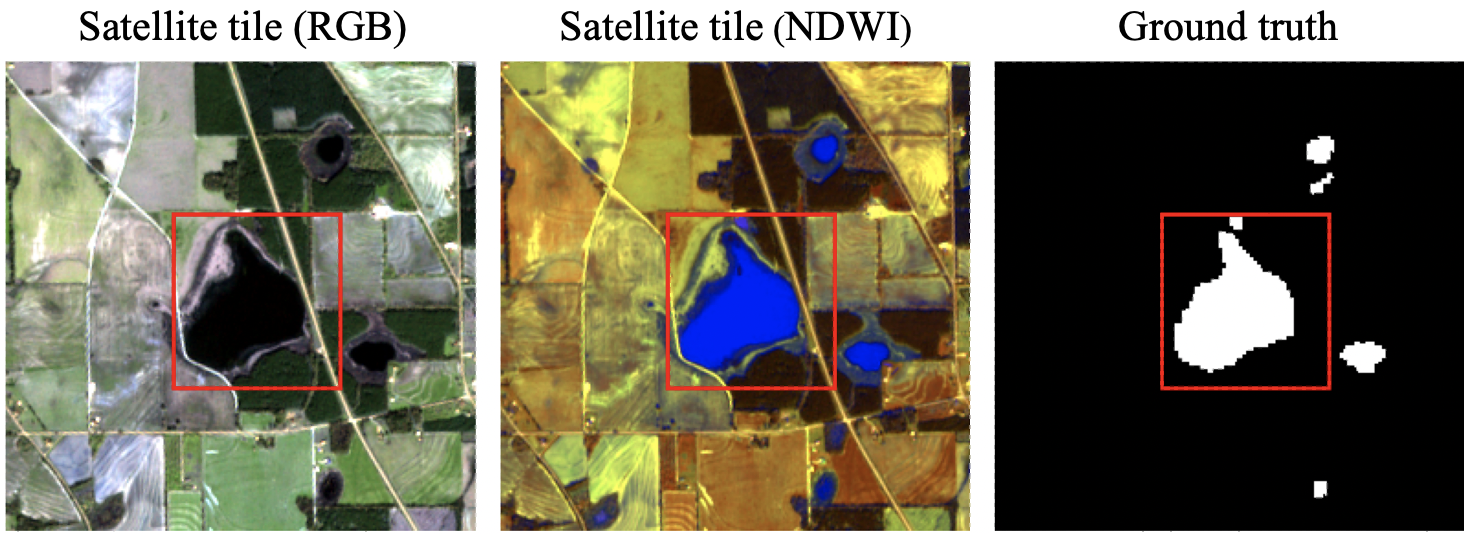}
	\caption{Example of a 256x256 satellite tile and corresponding ground truth flood mask from the WorldFloods dataset, with a bounding box.}
	\label{fig:TrainingData}
\end{figure}
\begin{itemize}
    \item Image pairs of Sentinel-2 data and corresponding labelled water masks are downloaded from HuggingFace \cite{Image_and_Signal_Processing;_ISP.}.
    \item  Data are cropped into tiles of 256x256 pixels, to facilitate identification of single water bodies. 
    \item Tiles containing clouds or not containing water bodies are removed.
    \item NDWI, the normalised difference water index, is calculated and added to the satellite data. NDWI increases the contrast between water bodies and vegetation.
    \item However, since SAM models have been pre-trained on 3-channel red, green and blue (RGB) data; they are not designed to handle 13-band satellite data. We therefore extract only the bands corresponding to the RGB channels (bands 2, 3 and 4). Our code also allows the replacement of one of the RGB bands with NDWI.
    \item The satellite data are normalised based on their standard deviation. 
    \item The ground truth data are converted to binary masks, where pixel values of 1 represent water.
    \item Finally, bounding box coordinates covering water bodies are extracted; SAM models require input prompts to guide the model during fine-tuning. The use of bounding boxes limits issues with class imbalance. 
\end{itemize}

After processing, the training subset contains significantly less data, only 1.3 GB (1840 tile pairs). The data is equally split among 8 satellites and is sufficiently small to be stored onboard the Unibap hardware. An important assumption is that labelled training samples are already available onboard the satellites. In the future we plan to rely on semi-supervised or self-supervised methods to compensate for this. 

\subsection{Segmentation Modelling}
\begin{figure}[!b]
    \centering
    \includegraphics[width=.99\columnwidth]{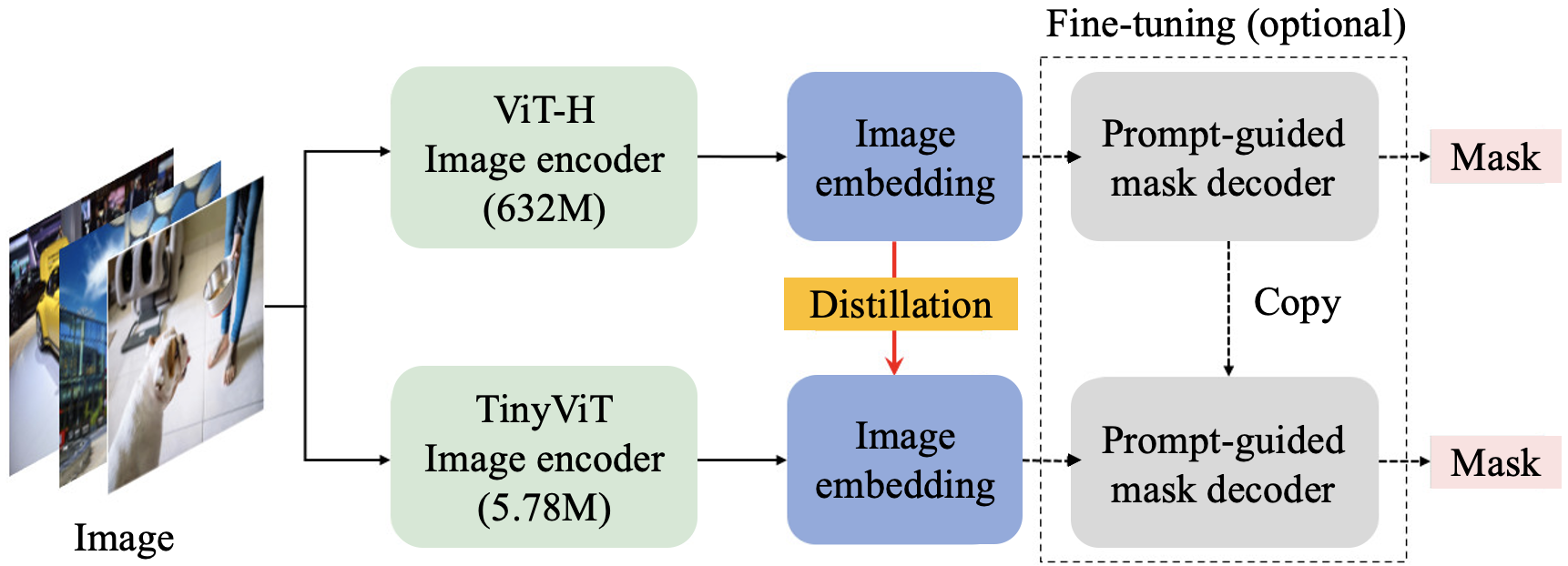}
	\caption{MobileSAM model architecture. Image adapted from \cite{Zhang_Han_Qiao_Kim_Bae_Lee_Hong_2023}.}
	\label{fig:MobileSAM}
\end{figure}

MobileSAM is a lightweight segmentation model designed to run on edge devices. The model architecture is shown in Fig. \ref{fig:MobileSAM}. Like the original SAM, it consists of an image encoder and mask decoder. However, MobileSAM is significantly smaller, containing 5.78 million parameters in the image encoder compared to 632 million in the original SAM. Both variants use the same mask decoder. 

In this work, we fine-tune only the mask decoder. The input images and bounding boxes are resized to 1024x1024 for model compatibility with the image encoder. While this is assumed to impact the memory footprint, modifications to the image encoder fall outside the scope of this preliminary investigation. We utilise sigmoid activation to handle the binary segmentation task and configure the code to allow easy variation of batch sizes, band combinations, learning rates, optimisers and loss functions. We identify the optimal configuration by maximising training and validation losses. The results presented use a batch size of 16, a learning rate of 1e-5, the Adam optimizer and mean squared error loss.

\subsection{Satellite Hardware}
To investigate the feasibility of onboard model fine-tuning, we benchmark this process on an iX10-100 satellite processor developed by Unibap. This radiation-tolerant processor is equipped with a diverse set of compute, including CPU, an onboard GPU and a VPU, to enable in-orbit processing of demanding space applications \cite{Unibap}. 

\subsection{Operational Constraints Modelling}
Running processes onboard satellite hardware will both impact and be impacted by the operational constraints of the satellite. Quantifying measurements such as the temperature of the hardware and the battery power is essential to determine the feasibility of running processes onboard. Additionally, in the context of rapid, distributed model fine-tuning, accurately modelling communication windows between satellites and the ground stations that they communicate with is critical. The PASEOS simulation software enables our evaluation of these aspects. More details about the software can be found in the related paper \cite{Gómez_Östman_Shreenath_Meoni_2023}.
\begin{figure}[!b]
    \centering
    \includegraphics[width=.99\columnwidth]{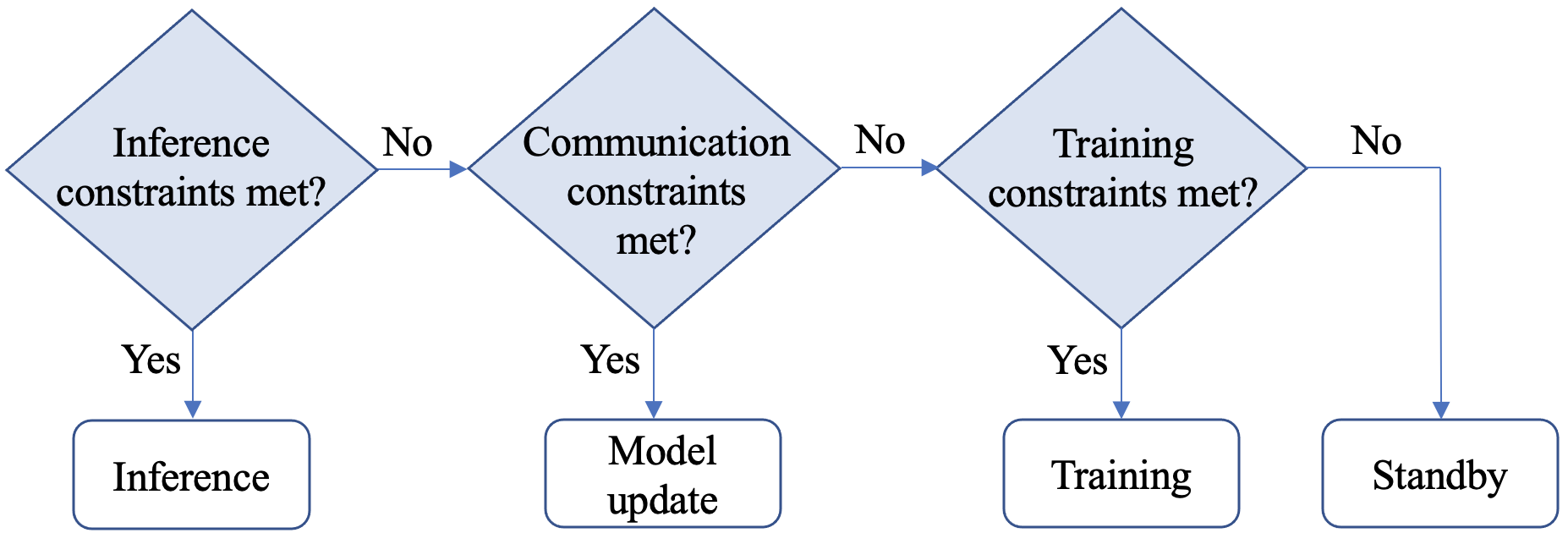}
	\caption{Flow chart showing the PASEOS decision-making process.}
	\label{fig:FlowChart}
\end{figure}

\begin{table}[!b]
\caption{Simulation Scenarios. Both scenarios contain 8 satellites in sun-synchronous orbits.}
\begin{center}
\begin{tabular}{|p{1.4cm}|p{2cm}|c|c|p{1cm}|}
\hline
\textbf{Simulation Scenario} & \textbf{Communication Agents} & \multicolumn{3}{|c|}{\textbf{Orbital Parameters}} \\
\cline{3-5}
& & \textbf{\textit{Altitude}} & \textbf{\textit{Inclination}} & \textbf{\textit{Orbits/day}} \\
\hline
1 & 3 ground stations & 786 km & 98.6 degrees & 14.3 \\
2 & 1 relay & 450 km & 97.4 degrees & 15.4 \\
\hline
\end{tabular}
\label{tabsims}
\end{center}
\end{table}
We configure simulation scenarios as described in Table.~\ref{tabsims}. In particular, we model the power consumption and thermal impact on satellites when performing the following operations: 1) fine-tuning the MobileSAM model on unique data that each satellite collects, 2) conducting inference with the MobileSAM model, 3) exchanging model weights when a satellite reaches a communication window. A simple flow-chart is shown in Fig. \ref{fig:FlowChart}. PASEOS is configured such that a satellite will stand-by and avoid partaking in one of the activities in Fig. \ref{fig:FlowChart} if the battery state-of-charge drops below $0.2$ or the temperature exceeds 40$^{\circ}$C. By exchanging models between the satellites, the data does not need to be transferred. Of particular note is the configuration of communication agents. In scenario 1, satellites within the constellation exchange their model parameters when they are within a communication window with one of three ground stations located on Earth. The ground stations are located at Matera (Italy), Svalbard (Norway) and Maspalomas (Spain) \cite{ESA}. In contrast, satellites in scenario 2 exchange their model updates via the European Data Relay Satellite System (EDRS), a geostationary relay satellite specifically configured for exchange of information between space-bound systems \cite{EDRS}. We subsequently evaluate how the frequency of model updates impacts the rate of model convergence during fine-tuning. In both scenarios, we assume a downlink speed of 10 Mbit/second when a satellite exchanges it's model with a ground station or the EDRS. Finally, to enable our modelling of a constellation of satellites, PASEOS is used with the message parsing interface (MPI) which we utilise to facilitate parallelisation. In this setup, each satellite is one MPI rank.

\section{Results}
\begin{figure}[!b]
    \centering
    \includegraphics[width=.99\columnwidth]{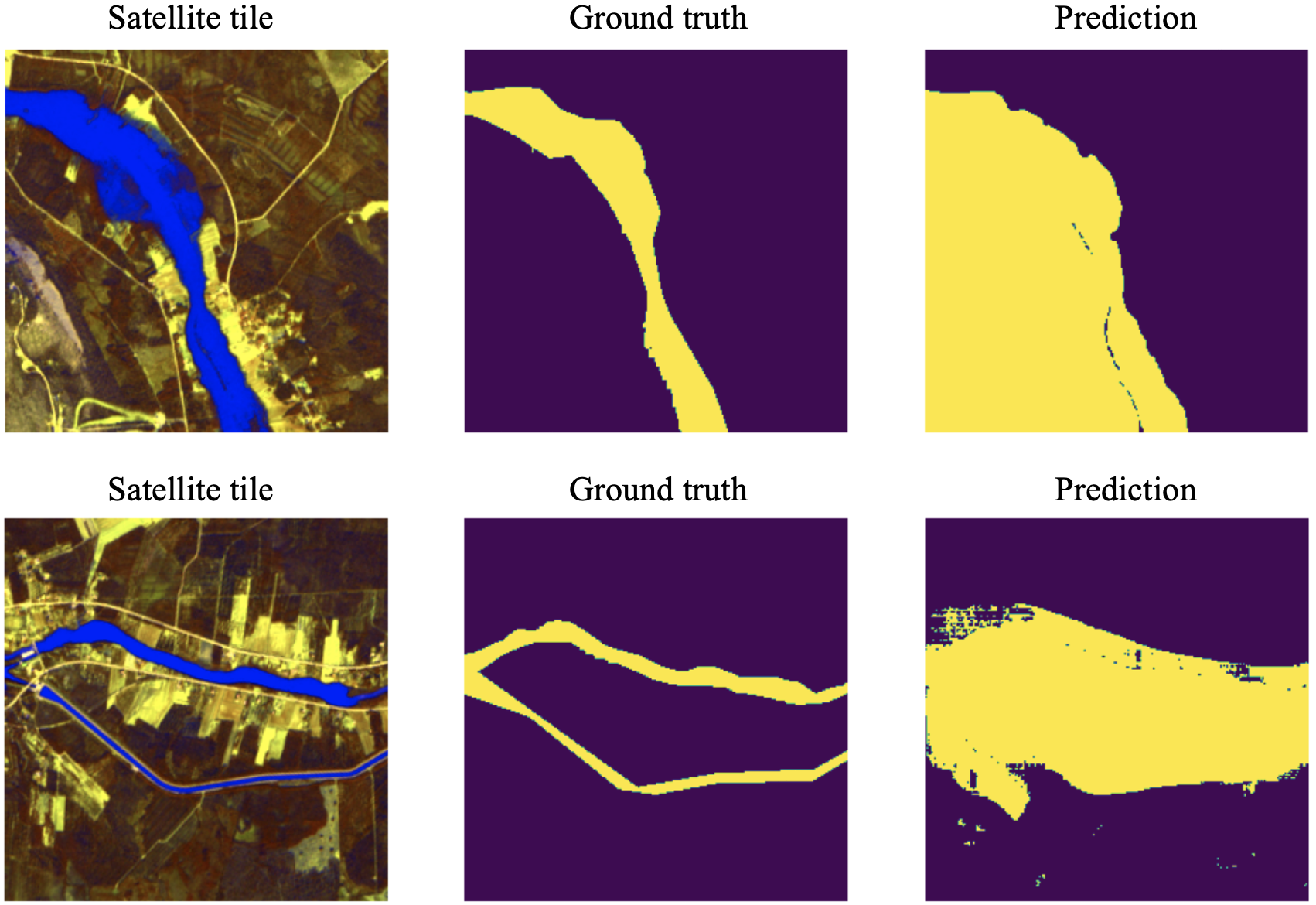}
	\caption{Segmentation performance of the MobileSAM model before fine-tuning. Images show water body samples from the disaster site test location at Ylitornio.}
	\label{fig:PreFT_NDWI}
\end{figure}

\begin{figure}[!t]
    \centering
    \includegraphics[width=.99\columnwidth]{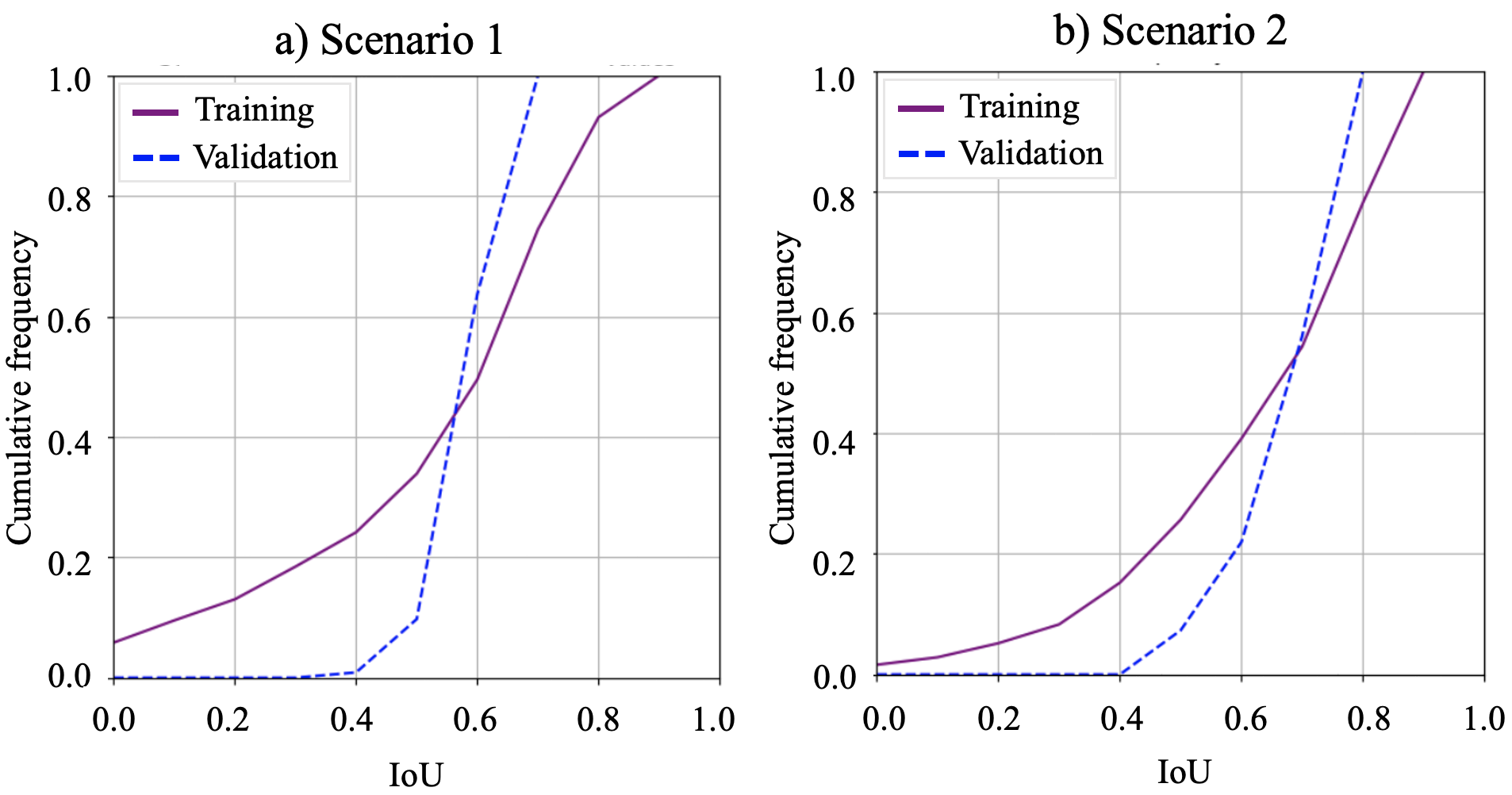}
	\caption{Cumulative frequency of IoU values for a satellite from each scenario.}
	\label{fig:IoUPlots}
\end{figure}

Before fine-tuning, baseline performance of the pre-trained MobileSAM model is evaluated, as shown in Fig. \ref{fig:PreFT_NDWI}. A disaster site from the original Worldfloods test dataset, located in Ylitornio, Finland, is selected. Ylitornio experienced flooding in 2018 \cite{copernicus}. The model has not been trained on any data from this region. From Fig. \ref{fig:PreFT_NDWI}, it can be seen that the model commonly overestimates the extent of a water body. The IoU before fine-tuning is measured at 0.47 on the 17 samples from the disaster site. Further IoU measurements on the full evaluation set of 167 tiles are displayed in Fig. \ref{fig:IoUPlots}.

The time required for model fine-tuning is benchmarked on the Unibap iX10-100 device; training one batch of 16 satellite and ground truth tile pairs takes 2.01 seconds on the Unibap GPU. Over a simulation period of 24-hours, the GPU can therefore fine-tune on 42,985 batches of data corresponding to 2,687 tile pairs. After benchmarking on the Unibap device, full simulations and model fine-tuning are conducted on NVIDIA A100-SXM4-40GB GPUs provided by the JASMIN data analysis facility \cite{jasmin_about}. The GPU cluster consists of 4 nodes each containing 4 GPUs, enabling our simulation of 8 satellites individually training on GPUs. 

\begin{figure}[!b]
    \centering
    \includegraphics[width=.99\columnwidth]{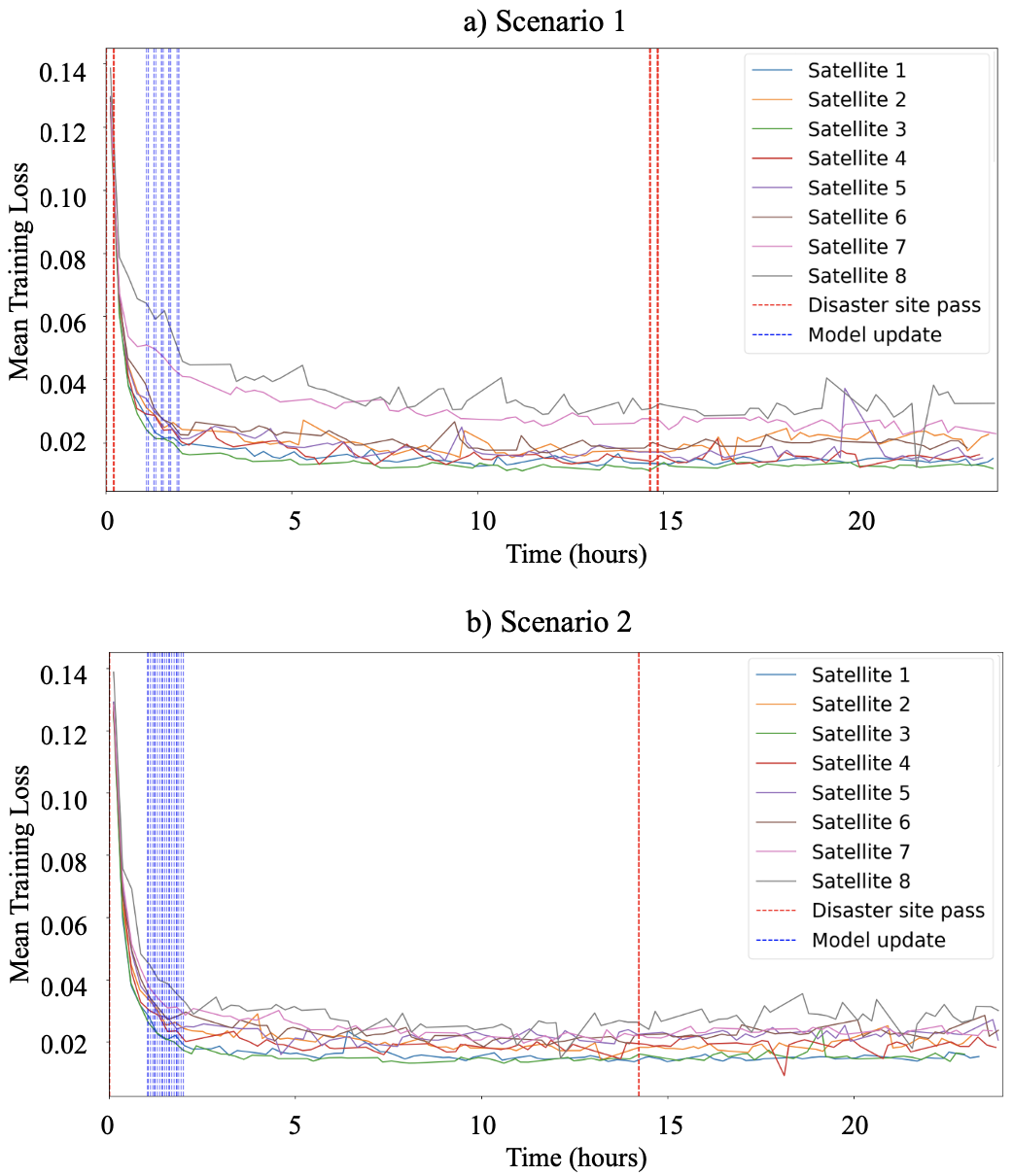}
	\caption{Satellite conditions during model training and evaluation. Model exchanges are shown for just a one hour period for demonstration purposes.}
	\label{fig:TrainingLosses}
\end{figure}

Since both simulation scenarios run for a 24-hour period, with a disaster site revisit time of nearly 15 hours, during this time at least one of the satellites within the constellation will pass over the disaster site. When any satellite in the constellation passes over the disaster site, the most recent fine-tuned model checkpoint is used to test segmentation performance. Fig. \ref{fig:TrainingLosses} shows the training loss over time for each scenario as the satellites train and exchange their model updates. In both simulation scenarios, the model appears to converge very rapidly, after just a few hours of training. Hence the satellites would be able to provide actionable information on the disaster much sooner. This could enable rapid alerting of disaster events within just a few hours of fine-tuning. If the same training were to be conducted on the ground, this would incur delays as each satellite first needs to transfer their data to a ground station.

Also of note is the frequency of communication updates. In scenario 1, satellites exchange their model updates via the ground stations 150 times. In contrast, satellites in scenario 2 exchange their model updates more frequently, 280 times, via the geostationary relay. From Fig. \ref{fig:TrainingLosses}, it can be seen that the training losses in both scenarios appear very similar, despite more frequent communication in scenario 2. The most notable difference is the more rapid convergence in scenario 2 within the first two hours. However, given the cost of using the EDRS for communication, this potentially outweighs the small observed benefit. These initial findings align with the work of \cite{ostman2023decentralised}, which provides more detailed analyses of different communication scenarios. This highlights the importance of scenario-specific optimisation, where communication frequency must balance convergence speed, model performance and operational expenses.

\begin{figure}[!b]
    \centering
    \includegraphics[width=.99\columnwidth]{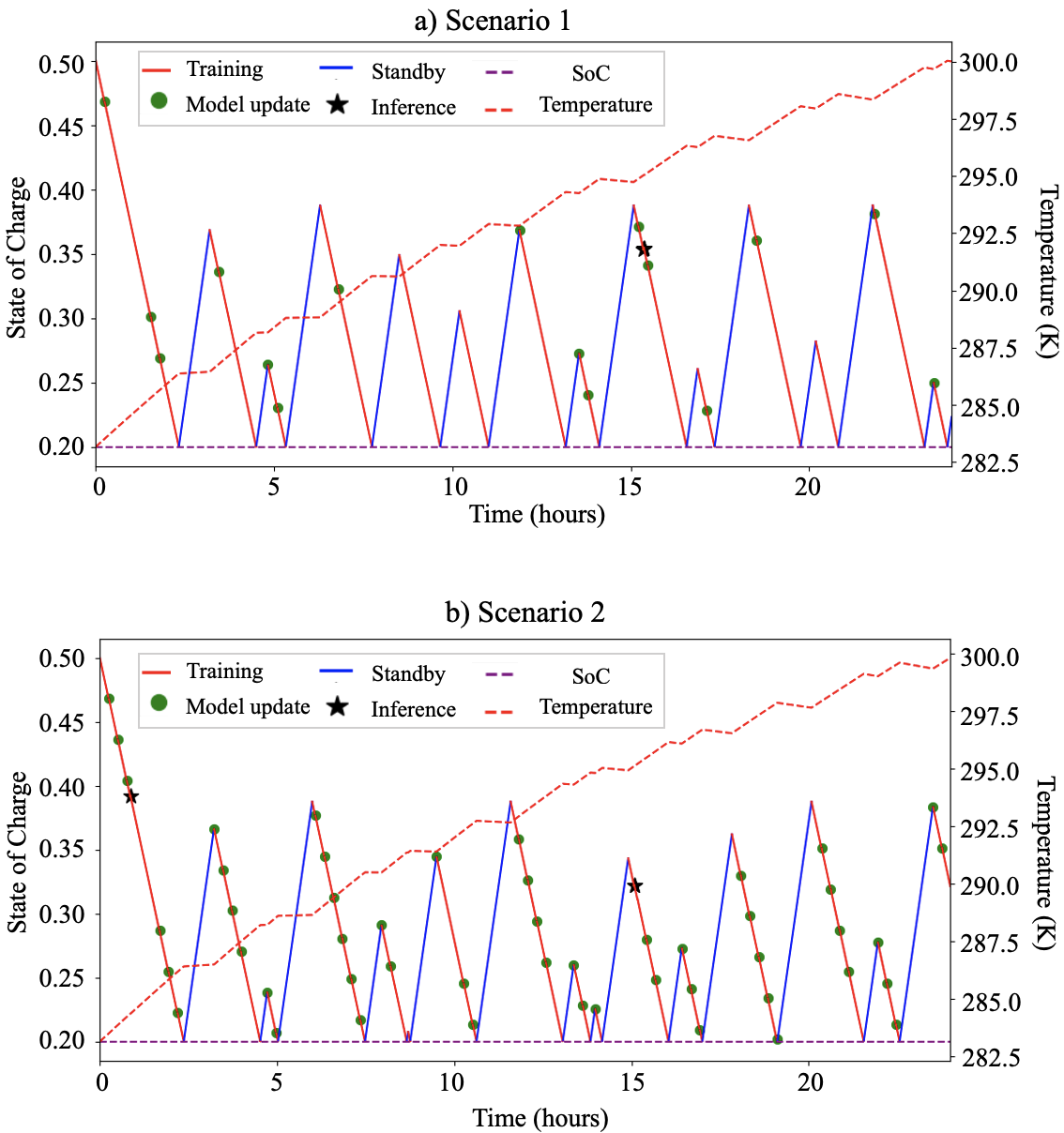}
	\caption{Example for one satellite from each scenario showing the state of charge over time as the satellite performs various activities.}
	\label{fig:SoC_1sat}
\end{figure}

Finally, parameters such as the state of charge and the temperature of the hardware are also captured during training and communication. An example of the state of charge is shown in Fig. \ref{fig:SoC_1sat} for one satellite in each scenario. The state of charge is colour-coded according to the activity the satellite was performing at the time. 

Fig. \ref{fig:SoC_1sat} also shows windows during which the satellite is in eclipse and hence not being charged by the sun. It can be observed that in both scenarios the satellites occasionally fall below their threshold state of charge. When this happens, the satellite remains in "Standby" mode until it is sufficiently recharged and can continue with fine-tuning.

\begin{figure}[!t]
    \centering
    \includegraphics[width=.99\columnwidth]{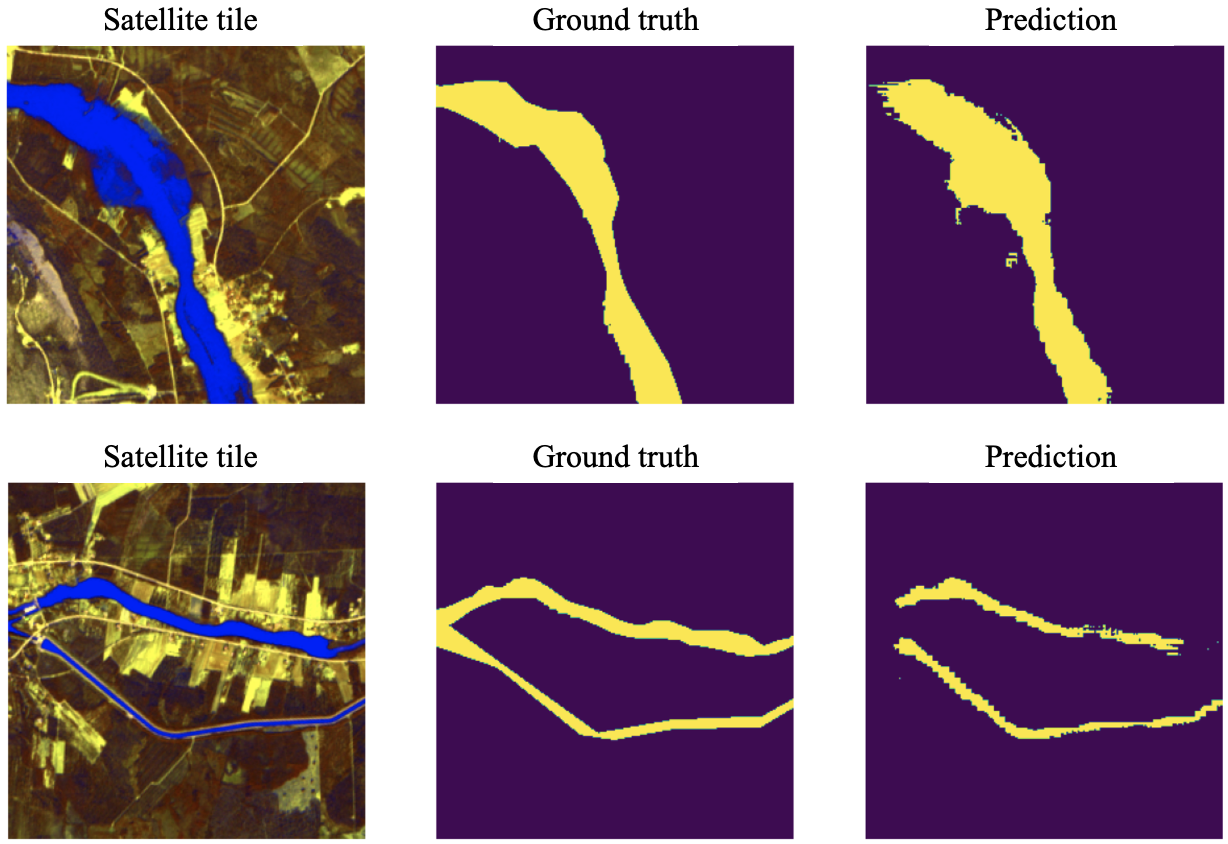}
	\caption{Segmentation performance of the MobileSAM model after fine-tuning under scenario 2. Images show water body samples from the disaster site test location at Ylitornio.}
	\label{fig:PostFT}
\end{figure}

In terms of segmentation performance at the disaster site, notable improvements are observed, as shown in Fig. \ref{fig:PostFT}. In some cases, where the ground truth is imperfect, the model results appear even superior to the ground truth. After fine-tuning with the NDWI band, the IoU score increases by 46.8\%, from 0.47 to 0.69 for scenario 2 and 36.2\% for scenario 1. We note that the reported IoU should be considered within the limitations of training on an imperfect ground truth dataset; a large number of ground truth samples within the WorldFloods dataset appeared to be incorrect. For example, some satellite images were entirely or partially covered by cloud, but not labelled as cloudy in the ground truth and hence missed in our filtering. Additionally, and more commonly, a water body may be labelled in the ground truth, but imperfectly. Samples which appeared obviously incorrect were removed from the filtered training dataset. However, it is still the case that most ground truth samples do not perfectly align with the water body in the satellite image, which will thus limit the training.

Despite very promising initial results, the fine-tuned model does struggle to perfectly map flooded regions, in contrast to clearly defined water bodies which it segments very well. Increasing the number of flood water examples in the training dataset is an immediate next step that can be explored and should yield fast improvements. Additionally, these scenarios can be repeated for other disaster sites to increase confidence in our measured improvement.

\section{Concluding Remarks}
Overall, our work demonstrates the rapid adaptability of MobileSAM to the EO domain when fine-tuned on a very small dataset. We also observe that, when accounting for operational constraints such as temperature and power, MobileSAM can effectively be fine-tuned onboard satellite hardware, benefiting from distributed learning within a satellite constellation and speeding up critical response time. These results are promising for the field of onboard EO data analysis; rapid fine-tuning can enable near-real-time analysis of hazards onboard satellite hardware, removing the bottlenecks associated with down-linking satellite data to ground stations. 

There are many avenues to explore in future studies. Firstly, more refined fine-tuning of the MobileSAM model can be explored by incorporating light-weight adaptor modules in the encoder. Such fine-tuning has recently been demonstrated with promising results by \cite{wu2023medical} and \cite{pu2024classwisesamadapter} for the original SAM.

Additionally, although the MobileSAM model demonstrates few-shot learning, requiring only a fraction of the original data for model fine-tuning, there still remains reliance on labeled training samples. Uploading training samples to a satellite is an additional bottleneck in terms of both latency and storage space and so this is a key limitation to address. Furthermore, overfitting risks associated with small, domain-specific datasets necessitate careful monitoring. Future work will consider self-supervised learning approaches to mitigate these potential biases and enhance model adaptability to diverse EO scenarios.

Finally, our fine-tuning uses Sentinel-2 data which has already undergone some ground-based processing. Subsequent work will investigate the efficacy of fine-tuning with the rawest forms of satellite data, as it would be acquired onboard.

\section{Supplementary}
\subsection{Data and Code Availability}
To facilitate reproducibility, code used for data acquisition and pre-processing, modelling and hardware simulation, are made available in the following repository: https://github.com/aidotse/Decentralized-EO/tree/main. Notebooks are also available for demonstration purposes.

\subsection{Conflicts of Interest}
The authors declare, to the best of their knowledge, no conflicting interests. 

\section*{Acknowledgments}
The authors thank the reviewers for their thoughtful comments and acknowledge Unibap, who provided access to the iX10-100 satellite processor. Meghan Plumridge acknowledges the UKRI Centre for Doctoral Training in the Application of Artificial Intelligence to the study of Environmental Risks (AI4ER), led by the University of Cambridge and the British Antarctic Survey. The work of Rasmus Maråk and Chiara Ceccobello was funded by the project SpaceEdge Phase 2 under Swedish National Space Agency grant P5-8-3-0003.

\end{document}